\documentclass[11pt, a4paper, logo, copyright, nonumbering]{kwaipilot}
\usepackage{dblfloatfix}
\usepackage{ulem}
\usepackage{caption}
\usepackage{dramatist}
\usepackage{xspace}
\usepackage{pifont} %
\usepackage{multirow}
\usepackage{tcolorbox}
\usepackage{xltabular}
\usepackage{longtable}
\usepackage{hyperref}
\usepackage{graphicx}
\usepackage{subcaption}
\usepackage{float}
\usepackage{amsfonts}
\usepackage{amsmath}
\usepackage{amssymb}
\usepackage{lineno}
\usepackage{multirow}
\usepackage{adjustbox}

\usepackage[bottom]{footmisc}

\usepackage{CJKutf8}
\usepackage{setspace}

\usepackage{dsfont}
\usepackage{array} %
\usepackage{tabularx} %
\usepackage{xcolor} %
\usepackage{tabularx}
\usepackage{booktabs}

\usepackage{lipsum}  %
\usepackage{multicol} %

\usepackage{indentfirst} 
\usepackage[sort&compress,numbers]{natbib}

\makeatletter
\def\@BTrule[#1]{%
  \ifx\longtable\undefined
    \let\@BTswitch\@BTnormal
  \else\ifx\hline\LT@hline
    \nobreak
    \let\@BTswitch\@BLTrule
  \else
     \let\@BTswitch\@BTnormal
  \fi\fi
  \global\@thisrulewidth=#1\relax
  \ifnum\@thisruleclass=\tw@\vskip\@aboverulesep\else
  \ifnum\@lastruleclass=\z@\vskip\@aboverulesep\else
  \ifnum\@lastruleclass=\@ne\vskip\doublerulesep\fi\fi\fi
  \@BTswitch}
\makeatother

\addto\extrasenglish{
}

 {\begin{list}{}%
         {\setlength{\leftmargin}{#1}}%
         \item[]%
 }
 {\end{list}}
 
\reportnumber{001} %
\renewcommand{\today}{}

\title{\centering SeamlessFlow: A Trainer–Agent Isolation RL Framework Achieving Bubble-Free Pipelines via Tag Scheduling}

\author{
\textbf{Jinghui Wang}$^{*,\dagger}$,
\textbf{Shaojie Wang}$^{*,\dagger}$,
\textbf{Yinghan Cui}$^{*}$,
\textbf{Xuxing Chen}$^{*}$,
\textbf{Chao Wang}$^{*}$,
\textbf{Xiaojiang Zhang}$^{*}$,
\textbf{Minglei Zhang}$^{*}$,
\textbf{Jiarong Zhang}$^{*}$,
Wenhao Zhuang,
Yuchen Cao,
Wankang Bao,
Haimo Li,
Zheng Lin,
Huiming Wang,
Haoyang Huang,
Zongxian Feng,
Zizheng Zhan,
Ken Deng,
Wen Xiang,
Huaixi Tang,
Kun Wu,
Mengtong Li,
Mengfei Xie,
Junyi Peng,
Haotian Zhang,
Bin Chen,
Bing Yu
}

\renewcommand{\phi}{\varphi}

\renewcommand{\epsilon}{\varepsilon}
\renewcommand{\imath}{\mathrm{i}}

\newlength{\restsubwidth}
\newlength{\restsubheight}
\newlength{\restsubmoreheight}
\newcommand{\rest}[2]{%
        \settowidth{\restsubwidth}{\ensuremath{#2}}
        \settoheight{\restsubheight}{\ensuremath{{}_{#2}}}
        \ensuremath{{#1\hskip 0.5pt}_{\vrule\kern2pt\parbox[b][%
        4pt][b]{\the\restsubwidth}{%
                        \ensuremath{{}_{#2}}}}}
        }

\begin{abstract}
We introduce SeamlessFlow, a server-based reinforcement learning (RL) framework that addresses two core challenges in industrial-scale RL: (1) decoupling RL training from the complex execution flow of agents; (2) maximizing GPU utilization with minimal idle time while preserving the stability and scalability required for large-scale deployments. First, SeamlessFlow introduces a data plane that decouples the RL trainer from diverse, complex agent implementations while sustaining high throughput. A central trajectory manager maintains complete interaction histories and supports partial rollout, allowing rollout to pause for weight updates and resume seamlessly, keeping agents unaware of service interruptions. Second, we propose a tag-driven scheduling paradigm that abstracts hardware into capability-tagged resources, unifying colocated and disaggregated architectures. Based on this, SeamlessFlow introduces a spatiotemporal multiplexing pipeline that dynamically reassigns idle training nodes to rollout in a train–rollout separated setup, eliminating pipeline bubbles and fully exploiting heterogeneous cluster resources. By combining these innovations, SeamlessFlow delivers both stability and high performance, making it well-suited for multi-agent, long-horizon, and other complex RL tasks.
\end{abstract}

\begin{document}

\maketitle



\section{Introduction}

In industrial-scale RL deployments, a single LLM serving layer typically backs multiple products and agents\cite{liu2024large,liuagentbench,yao2024tau}. Algorithm engineers who own RL post-training usually do not want to reason about each agent's internal workflow; they want a session-level history—ideally in token-id form—so that trajectories used for RL are bit-for-bit consistent with what the model actually saw during inference\cite{chen2023fireact, jin2025search, song2025r1}. This requirement becomes even more critical for agents with test-time-scaling\cite{zhang2025survey, snell2024scaling, muennighoff2025s1} or memory mechanisms\cite{chhikara2025mem0}, where one "session" can branch into multiple trajectories with different prefixes. At the same time, industrial RL must balance training stability with GPU throughput. To maximize GPU utilization, today's RL compute resource allocation strategies are organized around two architectural mindsets:
\begin{itemize}
    \item Colocated designs (e.g., VERL\cite{sheng2025hybridflow}, Kimi K2\cite{team2025kimik2}, Seed1.5-Thinking\cite{seed2025seed1}): colocate rollout and training on the same GPU resources with temporal multiplexing. This design minimizes pipeline bubbles and keeps GPUs highly utilized. However, it cannot fully exploit heterogeneous clusters, since every GPU must handle both rollout and training regardless of its suitability. Stability is also a concern—because rollout and training share the same resources, a crash in one serving process can cascade and bring down the entire RL training.

    \item Disaggregated designs (e.g., OpenRLHF\cite{hu2024openrlhf}, StreamRL\cite{zhong2025streamrl}, AReaL\cite{fu2025areal}, AsyncFlow\cite{han2025asyncflow}) place rollout and training on separate clusters, allowing stage-specific optimizations, flexible use of heterogeneous hardware, and uninterrupted serving—for example, new weights can be broadcast to one rollout partition while others continue generating. However, the separation inevitably induces pipeline bubbles due to stage dependencies. When rollout and training wall times drift (e.g. long chains of thought\cite{guo2025deepseek,wei2022chain} or agentic tasks), overlap techniques\cite{han2025asyncflow} cannot effectively eliminate the non-negligible pipeline bubble, significantly jeopardizing throughput.
\end{itemize}

These limitations are exacerbated in online RL scenario, where rollouts are collected through online applications or agents. In such case, we must simultaneously guarantee policy freshness, serving continuity, and high utilization across heterogeneous compute resources. 

To address these issues, we introduce SeamlessFlow, a server-based RL framework with two complementary innovations:

(1) Data plane for trainer–agent isolation with high throughput:
SeamlessFlow introduces a dedicated data plane that fully decouples the RL trainer from diverse and evolving agent implementations, while sustaining maximal throughput in multi-agent, long-horizon settings. At its core is a trajectory-manager service that sits between the LLM inference services and all downstream agents. The trajectory-manager service centrally captures every token-level input–response, reconstructing complete trajectories via longest-prefix matching over session histories—even when sessions branch due to memory mechanisms, test-time scaling, or other agent-side behaviors.

Rather than requiring each agent to implement precise rollout logging—which is both error-prone and costly to maintain—the trajectory-manager service becomes the single source of truth for RL training data. It provides all trajectories to the RL trainer in token-id form, ensuring bit-for-bit consistency between what the model generated during inference and what is consumed during training.

To maximize throughput, the data plane incorporates partial rollout\cite{team2025kimik1.5} seamlessly: when policy updates, or resource events occur, ongoing generations can be paused mid-turn, buffered, and later resumed without discarding previous work. The trajectory-manager service also makes such pause transparent to agents and avoids LLM inference services' interruptions. As a result, new agents can be added into RL trainer loop without impacting the RL pipeline, and RL engineers can operate independently of agent internals, achieving both stability and scalability in industrial deployments.

(2) Tag-driven scheduling for unified spatiotemporal resource allocation:
SeamlessFlow introduces a tagging system that abstracts hardware into capability-labeled resources, treating fully colocated and fully disaggregated deployments as special cases of a single scheduling paradigm. Each compute resource carries one or more capability tags indicating which tasks it can serve (e.g., rollout, update, critic, reward). From this perspective:
\begin{itemize}
\item Different tags across resources handle heterogeneity, mapping each task to the most suitable hardware.
\item Multiple tags on the same resource enable spatiotemporal multiplexing, removing pipeline bubbles by allowing instant role switching without waiting for the next training cycle.
\end{itemize}

By assigning different tags to different machines while also allowing certain machines to carry multiple tags, SeamlessFlow combines the high utilization of colocated setups with the stability and heterogeneity benefits of disaggregated designs.
In this setup, SeamlessFlow leverages the temporal multiplexing characteristic within the spatial multiplexing disaggregated architecture to reassign idle training resources to rollout tasks in real time, completely eliminating pipeline bubbles that typically persist in disaggregated systems.
As a result, this design supports online RL without risking interruptions to LLM services, and maintains high throughput in heterogeneous clusters.

In summary, our contributions are:
\begin{itemize}
    \item \textbf{SeamlessFlow framework.} A server-based RL framework for industrial-scale, multi-agent, long-horizon scenarios, sustaining high throughput on heterogeneous clusters.
    \item \textbf{Data plane for trainer–agent isolation.} A service-based layer that decouples training from diverse agents, reconstructs complete token-level histories via longest-prefix matching, and supports transparent partial rollouts to enable policy updates without disrupting generation.
    \item \textbf{Tag-driven scheduling.} A unified spatiotemporal resource allocation paradigm where capability tags map tasks to suitable hardware and allow instant role switching, combining colocated efficiency with disaggregated stability and minimizing pipeline bubbles.
\end{itemize}

\section{System Overview}
\begin{figure}[h]
\centering
\begin{minipage}{0.95\textwidth}
    \centering
    \includegraphics[width=\textwidth]{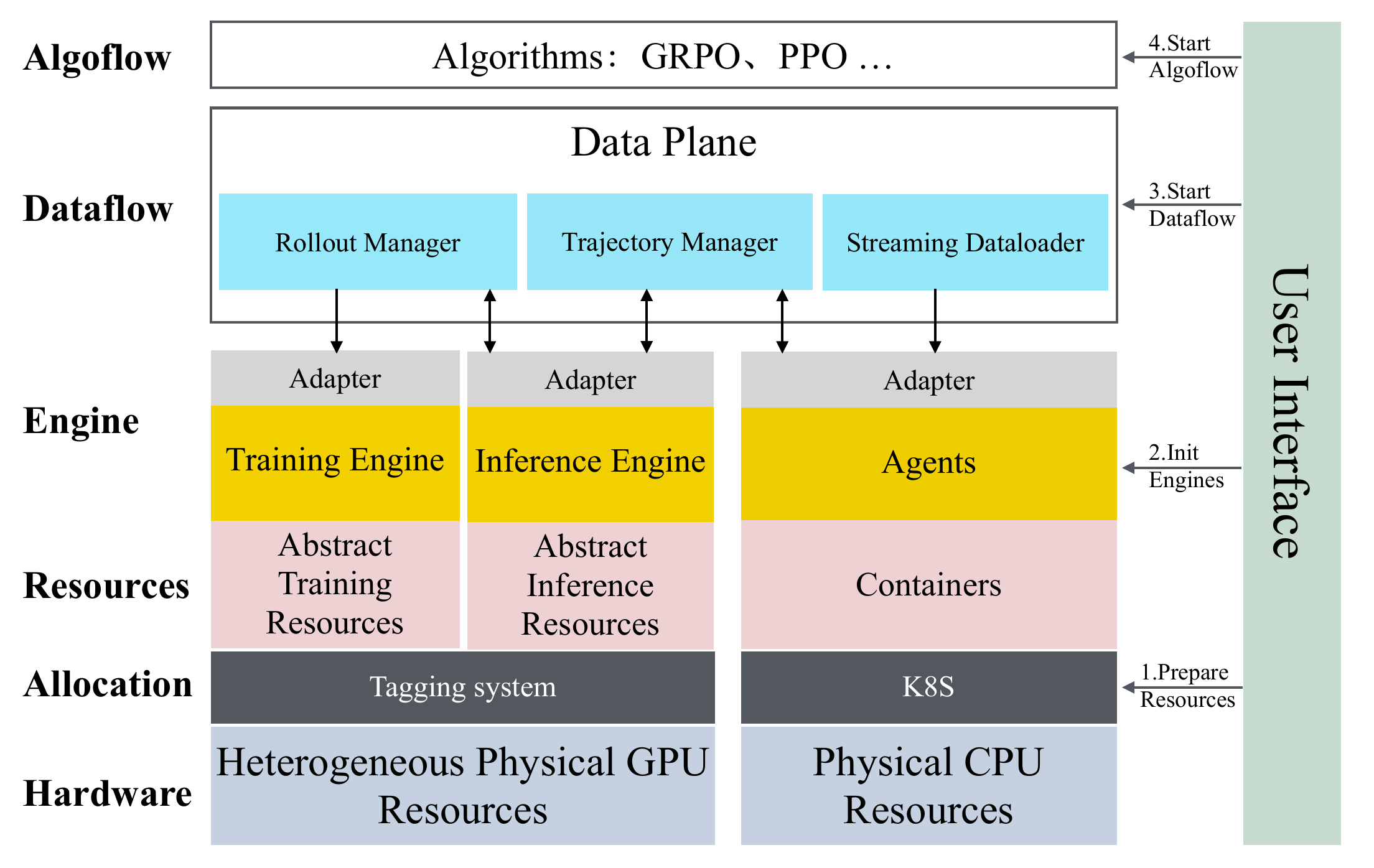}
\end{minipage}
\caption{\centering Hierarchical Design of SeamlessFlow.}
\label{fig:system_overview}
\end{figure}
As shown in Figure~\ref{fig:system_overview}, SeamlessFlow adopts a seven-layer architecture that spans from raw compute to the RL algorithm layer, with each stage designed to address the scalability, heterogeneity, and throughput demands of industrial-scale multi-agent online RL.

At the base, heterogeneous physical GPU and CPU resources provide the raw compute capacity. Above this, SeamlessFlow implements a tag-driven resource allocation layer—an innovation over traditional colocated or disaggregated architectures. Each device is assigned one or more capability tags (e.g., rollout, update, critic, reference), allowing tasks to be matched to the most suitable hardware and enabling task switching on a single device without waiting for a new training cycle. Tags can be reassigned on-the-fly in response to workload changes or failures, and scheduling operates at the tag-pool level rather than binding roles to fixed machines. This enables the combination of the high throughput characteristic of colocated setups with the flexibility and resilience of disaggregated systems.

On top of the Allocation layer, abstract training resources and abstract inference resources decouple hardware from the engines that consume them. These resources are bound to the training engine(e.g., Megatron-LM\cite{shoeybi2019megatron}, FSDP\cite{zhao2023pytorch}), inference engine(e.g., vLLM\cite{kwon2023efficient}, SGLang\cite{zheng2024sglang}), and agent containers respectively, enabling independent optimization of training and rollout while still allowing partial hardware sharing for utilization gains.

The fifth layer, Data Plane, is the other major innovation. It comprises the Trajectory Manager, Rollout Manager, and Streaming Dataloader, each implemented as an independent service. These services together enforce trainer-agent isolation while sustaining high throughput. The Trajectory Manager records every LLM input–output pair, reconstructs full sequences via longest-prefix matching, and guarantees trajectory consistency—ensuring that the sequences used for RL training are bit-for-bit identical to those observed during serving. The Rollout Manager buffers and resumes interrupted generations transparently, making weight updates and resource preemptions invisible to agents, while the Streaming Dataloader continuously dispatches tasks to agents, avoiding idle time caused by fixed batch scheduling.

At the algorithm layer, standard RL algorithms (e.g., GRPO\cite{shao2024deepseekmath}, PPO\cite{schulman2017proximal}) consume trajectories from the Data Plane without any dependency on agent internals. In practical use, the user has clear control over every layer for customization, including the algorithm, dataflow scheduling, tagging policy, and so on. Training can proceed in the following steps: (1) prepare resources, (2) initialize engines, (3) start dataflow, and (4) start the RL algorithm loop, as illustrated in Figure~\ref{fig:system_overview}.

\section{Data Plane: Trainer-Agent Isolation and High-Throughput Pipeline}
\label{gen_inst}
\begin{figure}[h]
\centering
\begin{minipage}{0.95\textwidth}
    \centering
    \includegraphics[width=\textwidth]{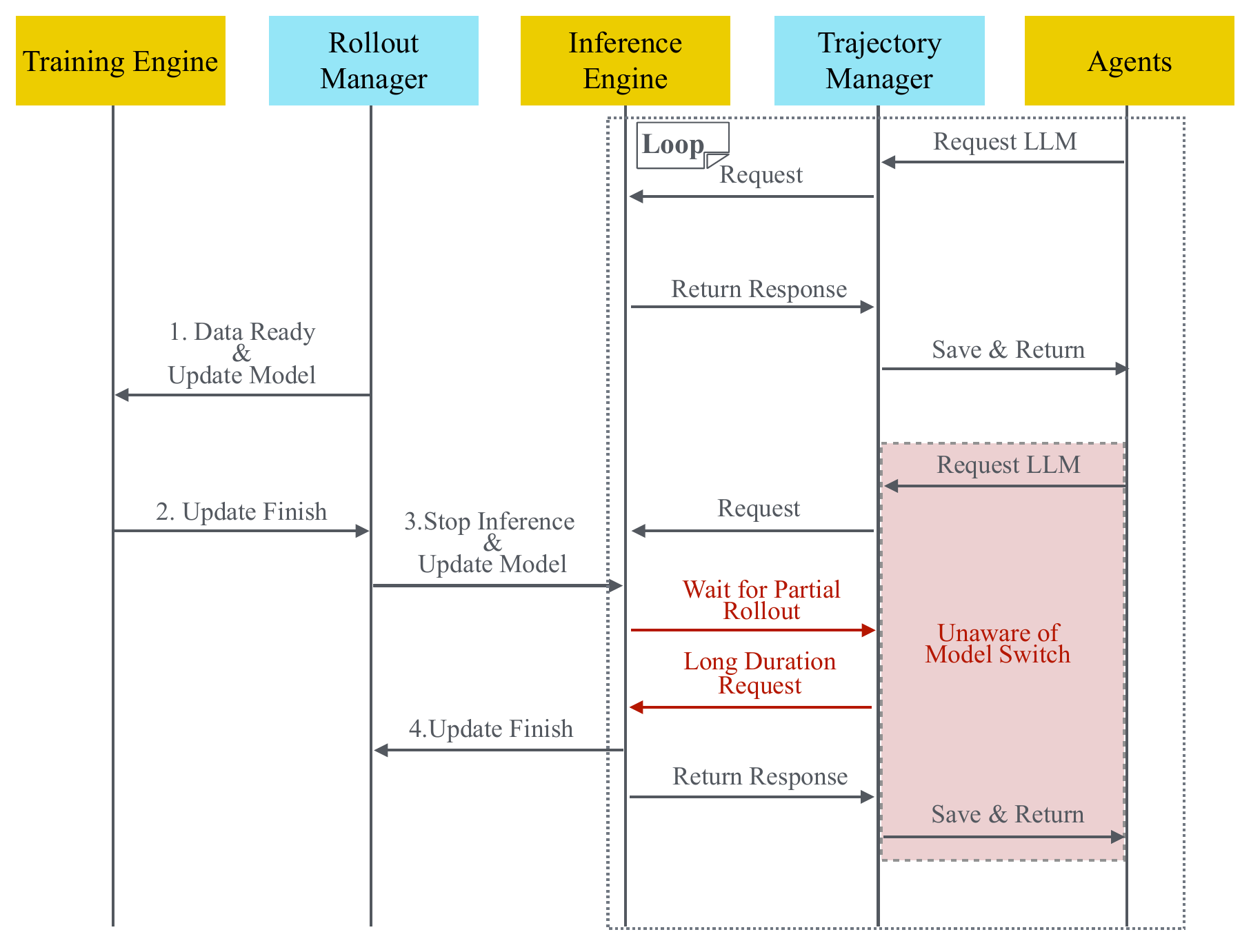}
\end{minipage}
\captionsetup{justification=raggedright,singlelinecheck=false}
\caption{Sequence diagram of the Rollout Manager and Trajectory Manager in the Data Plane, showing how the Trajectory Manager makes the agent oblivious to model update and how the Rollout Manager controls the switching between training and rollout phases.}
\label{fig:traj_manager}
\end{figure}
In industrial-scale RL for LLM-based agents, it is not desirable for the RL trainer to be tightly coupled to the diverse and evolving implementations of agents. Different agents vary in architecture, memory mechanisms, and auxiliary pipelines, and asking each to implement training-compatible logging, rollout control, or policy tracking would be both error-prone and costly to maintain. To sustain high throughput in such environments, partial rollout is essential: ongoing generations can be paused mid-turn when batch-size thresholds, policy weight updates, or resource events occur, and later resumed without wasting completed work. Existing trainer-agent-isolation frameworks\cite{luo2025agent} generally ignore partial rollout handling, meaning that weight updates during inference can leak to agents in the form of truncated or inconsistent responses.

As shown in Figure~\ref{fig:system_overview}, SeamlessFlow addresses these challenges through a dedicated data plane that achieves trainer-agent isolation with high-throughput pipelines. This data plane consists of three key modules: a trajectory manager, which makes agents and trainer oblivious to each other, captures all token-level I/O from LLM inference services and ensures trajectory consistency; a rollout manager, which controls the triggers to pause or resume rollouts, manages policy weight updates and other signals for training; and a streaming dataloader, which continuously dispatches tasks to agents to keep rollouts saturated. 

\paragraph{Rollout Manager}  
In SeamlessFlow’s data plane, the Rollout Manager is responsible for both determining when to interrupt ongoing rollouts and ensuring that paused rollouts are resumed seamlessly when conditions permit.  

\textbf{Interruption triggers} include:
\begin{itemize}
    \item Samples reaching the predefined threshold (e.g., batch size, timeout).
    \item Synchronization of new policy's weights.
    \item Tag transitions (e.g., a rollout resource reassigned as an actor). Details are provided in the next section.
\end{itemize}

\textbf{Resumption triggers} include:
\begin{itemize}
    \item Tag transitions (e.g., a node switched from actor back to rollout).
    \item Resource restoration events (e.g., a new resource added to the pool).
\end{itemize}

\paragraph{Trajectory Manager} The Trajectory Manager is the key module enabling trainer–agent isolation in SeamlessFlow.  
It acts as a proxy service inserted between agents and the LLM serving engine: all agents' requests to the LLM service have to pass through the Trajectory Manager service before reaching the LLM serving service, and all responses from LLM's responses pass back through it to the agents.  
An adapter layer on the LLM serving engine ensures that for each request, both the input and output are returned to the Trajectory Manager, which records them before forwarding the model output to agent.

A key requirement of this design is trajectory consistency: each trajectory must be an exact token-level replica of the output the LLM generated during the agent’s execution in the same task. 
This consistency is critical for accurate log-probability computation and stable credit assignment, particularly in multi-agent and long-horizon scenarios where a single session may branch into multiple prefixes.

Because the trajectory manager serves as the single source of truth for all RL data, it must handle significant storage pressure. To mitigate this, SeamlessFlow employs longest-prefix matching (LPM) to merge multi-turn dialogues belonging to the same session when they share identical prefixes. This merging builds a prefix tree for each session, eliminating redundant storage and avoiding recomputation of logprobs for shared prefixes, thereby improving both memory efficiency and computational throughput of the RL trainer.

In RL training, model weight updates can interrupt LLM serving, potentially exposing agents to inconsistent responses. To keep agents oblivious to such interruptions, the trajectory manager must handle these events transparently and preserve seamless generation across model switches.
To achieve this, when the Rollout Manager triggers a model update, the inference engine pauses, reloads the updated model, and then resumes generation. Any incoming requests during this phase receive a wait signal from the inference engine, upon which the Trajectory Manager issues a long-duration request that remains pending until the model switch completes. The inference engine then continues generation from the previously produced tokens using the new model, and the resulting output is recorded by the Trajectory Manager before being returned to the agents. Details are illustrated in Figure~\ref{fig:traj_manager}

Additionally, due to the partial rollout mechanism, a single trajectory may contain tokens generated by different model versions.  
To enable precise on-/off-policy separation, the trajectory manager annotates each token with the version of the model that generates it.

\paragraph{Streaming Dataloader} The streaming dataloader continuously feeds tasks to agents in a flow-based manner rather than waiting for fixed-size batches. By dispatching new requests as soon as resources are available, it keeps the rollout pipeline saturated, avoids idle gaps between tasks, and sustains higher overall throughput.

Together, these components form a producer–consumer architecture: from the trainer’s perspective, the trajectory manager acts as the producer of token-aligned trajectories, while the rollout manager consumes these trajectories to trigger model updates; from the agent’s perspective, the streaming dataloader produces tasks, and the trajectory manager consumes the resulting trajectories.
This bidirectional producer–consumer flow overlaps rollout with training, hides preemptions from agents, and preserving strict training–inference consistency.

\section{Tagging System: Unified spatiotemporal scheduling paradigm}

\subsection{Motivation}
Industrial-scale RL clusters must reconcile two often conflicting demands: 
(1) high compute resources utilization, which favors colocated architectures that multiplex rollout and training on the same devices without pipeline bubbles; 
(2) stability and heterogeneity accommodation, which favors disaggregated architectures that cleanly separate stages and match tasks to the most suitable hardware.  
However, both fully colocated and fully disaggregated setups limit flexibility: fully colocated setups suffer from stability issues and poor adaptation to heterogeneous hardware, while fully disaggregated setups incur pipeline bubbles when stage latencies drift. We therefore seek a unified scheduling abstraction that preserves the efficiency of colocated architectures while retaining the modularity and stability of disaggregated designs.

\subsection{Core Concept: Tag-Driven Scheduling}
SeamlessFlow introduces a tag-driven mechanism that decouples resource scheduling from physical machine assignments.  
Each compute node in the resource pool maintains two types of tags:  
\begin{enumerate}
    \item a capability tag describing which role in RL the resource can serve (e.g., \texttt{rollout} for rollout phase; \texttt{train} for policy updates, logprob computation, critic model's forwards and updating);  
    \item an active tag indicating the task the resource is currently executing.  
\end{enumerate}

Once capability tags are assigned, all physical resources are transformed into abstract resources—logical units grouped by their capabilities rather than their physical identity.  
The collection of all such abstract resources forms the abstract resource pool, which the scheduler queries to locate resources suitable for a given task type.

At runtime, the scheduler issues tasks by querying the abstract resource pool for resources carrying the required capability tags, and dispatches tasks to them in an SPMD manner.  
The scheduler supports preemptive allocation: it may assign a new task not only to idle resources, but also to currently occupied ones, in which case the ongoing task is interrupted and replaced with the new assignment.  
Preemption is essential for temporal multiplexing—when the same machine switches from one task to another, the previous task must be stopped before the new one starts. Fully colocated architectures inherently rely on such preemption to switch rollout and training phases. 

\subsection{Spatiotemporal Multiplexing Pipeline}

From the tagging viewpoint:  
\begin{itemize}
    \item \textbf{Fully colocated:} every machine carries all relevant tags (\{\texttt{rollout}, \texttt{train}, \ldots\}), allowing any device to serve any role at any time. This maximizes instantaneous utilization but risks instability and is not feasible for hetergeneous clusters.
    \item \textbf{Fully Disaggregated:} each machine carries exactly one tag, dedicating it to a single training stage. This maximizes specialization and stability but introduces pipeline bubbles.
\end{itemize}

Both extremes are special cases of tag-driven scheduling with tag assignments.  
The tagging perspective generalizes beyond these by adopting flexible tag configurations that enable partial time-sharing of resources.  
To achieve both high throughput and system stability, we further provide a tag allocation strategy for a Spatiotemporal Multiplexing Pipeline: 
\begin{enumerate}
    \item At initialization, SeamlessFlow assigns: (1) a subset of machines with both \texttt{rollout} and \texttt{train} capability tags; (2) the remaining machines with only the \texttt{rollout} capability tag.
    \item During the first rollout step, the scheduler observes that all resources are idle and that all resources carry the \texttt{rollout} tag. Consequently, all machines perform rollout in this phase.
    \item When the rollout manager detects that enough trajectories have been generated to perform an update, the task scheduler queries the resource pool for machines carrying the \texttt{train} tag.  
    \item Those machines—whether idle or currently executing rollout—are preempted and reassigned to the training task (If additional machines carrying the \texttt{train} tag have joined the cluster before this point and are currently idle, they are directly assigned the update task without preemption).
Since some resources in the cluster carry only a single \texttt{rollout} capability tag, these resources are not interrupted and continue executing rollout. 
As a result, from this moment onward, rollout and training proceed concurrently: the rollout stream continues to produce data, but relative to the most recent policy update, this data is now off-policy by one step lag.

    \item Once the training phase completes, the scheduler again assigns all machines with the \texttt{rollout} capability tag back to rollout tasks, restoring full rollout capacity. 
\end{enumerate}

The above process repeats for subsequent steps, enabling a high-throughput pipeline and achieving almost no idle compute resources.  

\begin{figure}[h]
\centering
\begin{minipage}{0.95\textwidth}
    \centering
    \includegraphics[width=\textwidth]{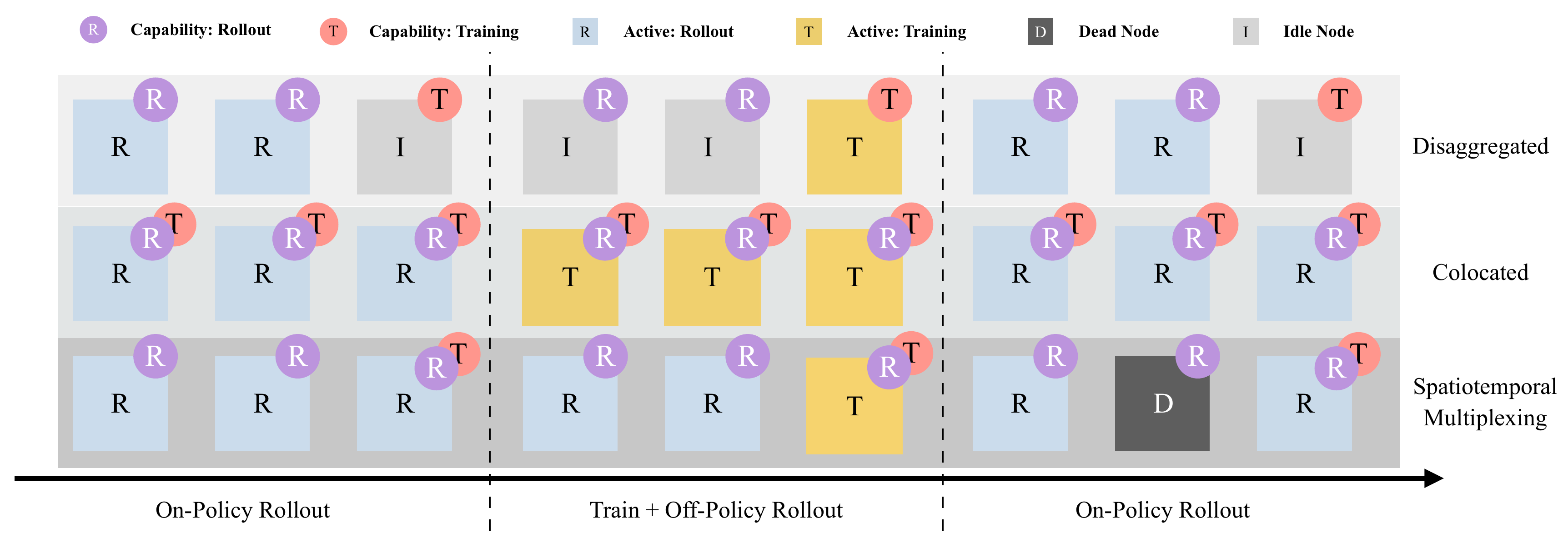}
\end{minipage}
\captionsetup{justification=raggedright,singlelinecheck=false}
\caption{Disaggregated, colocated and the proposed spatiotemporal multiplexing pipeline, from the tagging perspective. Note that under the preemptive allocation mechanism, our Tag-Driven system inherently tolerates node failures.}
\label{fig:Tag-driven system}
\end{figure}

As shown in Figure~\ref{fig:Tag-driven system}, fully disaggregated architectures inevitably suffer from pipeline bubbles: because RL stages execute sequentially, machines assigned to a later stage remain idle while earlier stages are still running.  
In contrast, the tag-driven approach can dynamically reassign such idle machines to do rollout tasks during these gaps, converting otherwise wasted capacity into productive work and effectively eliminating bubbles. 

Beyond bubble elimination, heterogeneous clusters require careful tag placement. For instance, assigning both \texttt{rollout} and \texttt{train} tags to compute-bound devices can bottleneck model training stages, harming overall throughput. To address this, we introduce a \texttt{train\_priority} tag alongside \texttt{rollout} and \texttt{train} tags. Higher \texttt{train\_priority} indicates a device is better suited for dynamically switching to training. We determine this priority via the classical roofline model \cite{williams2009roofline}, jointly considering each device’s HBM bandwidth and peak compute performance to guide optimal tag assignment in heterogeneous environments.

\begin{figure}[h]
\centering
\begin{minipage}{1.05\textwidth}
    \centering
    \includegraphics[width=\textwidth]{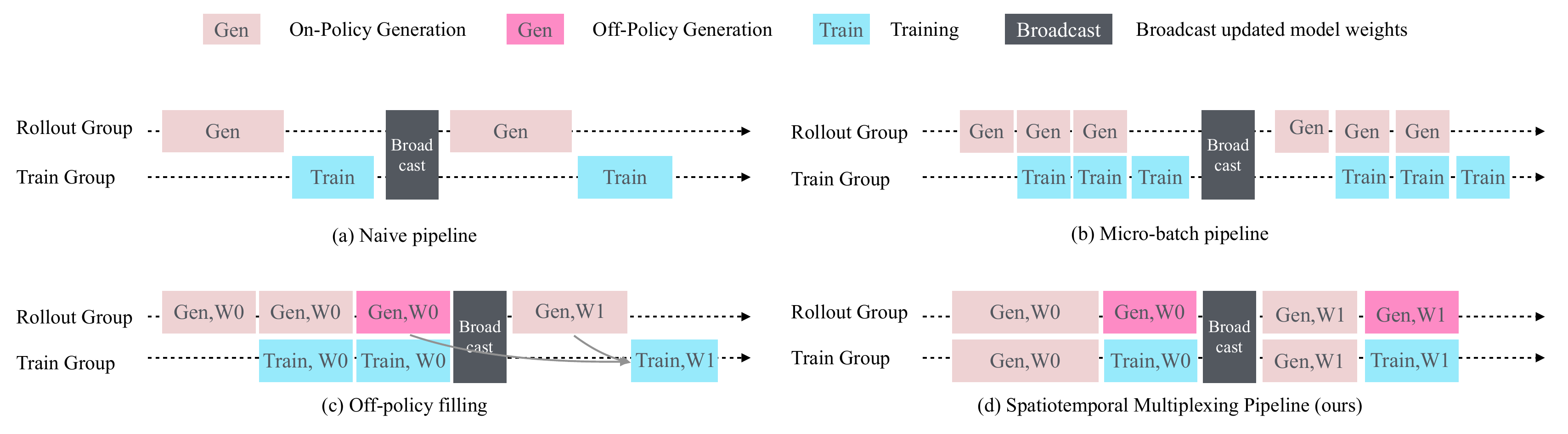}
\end{minipage}
\caption{\centering Illustration of multiple pipeline implementations.}
\label{fig:pipeline}
\end{figure}

\subsection{Pipeline Bubble Analysis} 
As shown in Figure~\ref{fig:pipeline}, we further illustrate and compare multiple pipeline implementations under the disaggregated setup, analyzing the pipeline bubbles in each design.
\begin{itemize}
    \item Naive Pipeline: Rollout and training are performed one after the other without overlap. Since the next stage can only start after the previous stage finishes, large idle periods appear between stages.
    \item Micro-batch Pipeline\cite{luo2025deepcoder}: Each batch is split into smaller micro-batches, and the gradient calculation begins as soon as a micro-batch is ready. This reduces bubbles but still leaves part of the rollout cluster idle in every step, because all samples have to remain strictly on-policy.
    \item Off-policy filling \cite{luo2025deepcoder,noukhovitch2024asynchronous,wang2024distrl,han2025asyncflow}: Off-policy data is used to fill rollout idle time of the Micro-batch Pipeline. However, in agentic RL, rollout often takes much longer than training, and performance depends on keeping the data as on-policy as possible. Under these conditions, the training cluster becomes chronically data-starved and idles while waiting for rollout output.
    \item Spatiotemporal Multiplexing Pipeline (ours) – Temporal multiplexing is introduced into a disaggregated setup: when training nodes are idle, they are reassigned to rollout tasks. This removes the rigid separation between "training cluster" and "rollout cluster", converting idle compute into productive work and achieving near bubble-free utilization.
\end{itemize}

\section{Performance}
To validate the superiority of SeamlessFlow, we conducted several experiments from two dimensions: 1) performance comparison in typical RL training scenarios; 2) effectiveness in downstream agentic tasks. The results demonstrate that SeamlessFlow not only exhibits advantages in training throughput but also achieves remarkable performance improvements in complex agentic tasks.

\subsection{Training Efficiency}
We conducted throughput comparisons with VERL, the mainstream reinforcement learning training framework. The comparisons adopted identical reinforcement learning algorithms, training datasets, hyperparameter settings (including learning rates, batch sizes, optimizer parameters, etc.). All experiments were executed on H800 GPU clusters of the same specifications and quantities.

\begin{figure}[h]
\centering
\begin{minipage}{0.9\textwidth}
    \centering
    \includegraphics[width=\textwidth]{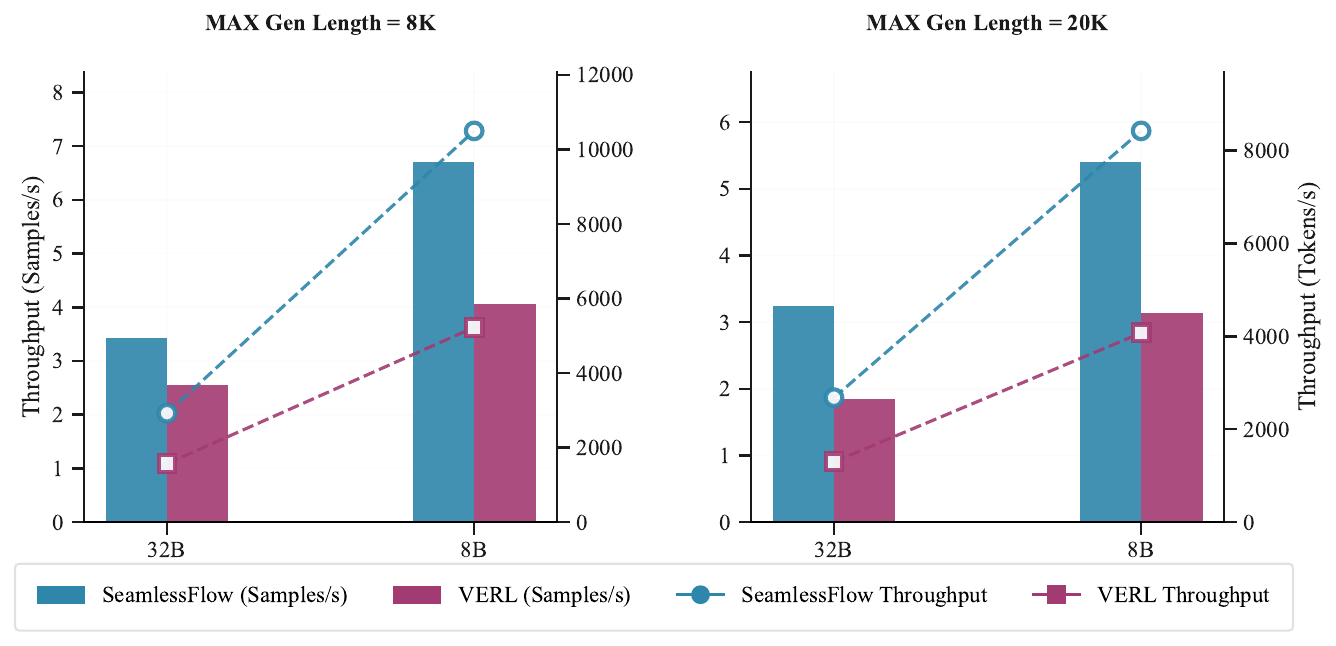}
\end{minipage}
\captionsetup{justification=raggedright,singlelinecheck=false}
\caption{Throughput comparison between SeamlessFlow and VERL frameworks. Left: Sample-level throughput (samples/second) for Qwen3-8B and Qwen3-32B models with maximum sequence length of 8k tokens. Right: Token-level throughput (tokens/second) with maximum sequence length of 20k tokens.}
\label{fig:Throughput}
\end{figure}

\subsubsection{Single-turn RL Performance}
For single-turn experiments, we compare the performance of SeamlessFlow and VERL frameworks on math tasks using a cluster of 32 H800 GPUs. As demonstrated in Figure \ref{fig:Throughput}, SeamlessFlow achieves an 100\% improvement in token throughput compared to VERL on average, resulting in a 62\% reduction in overall training time. These performance gain primarily stems from SeamlessFlow's dedicated data plane layer, which streams tasks in a producer–consumer manner, enabling continuous data flow between rollout and training without idle periods.

    

\subsubsection{Agentic RL Performance}
\begin{figure}[h]
\centering
\begin{minipage}[b]{0.48\textwidth}
  \centering
  \includegraphics[width=\linewidth]{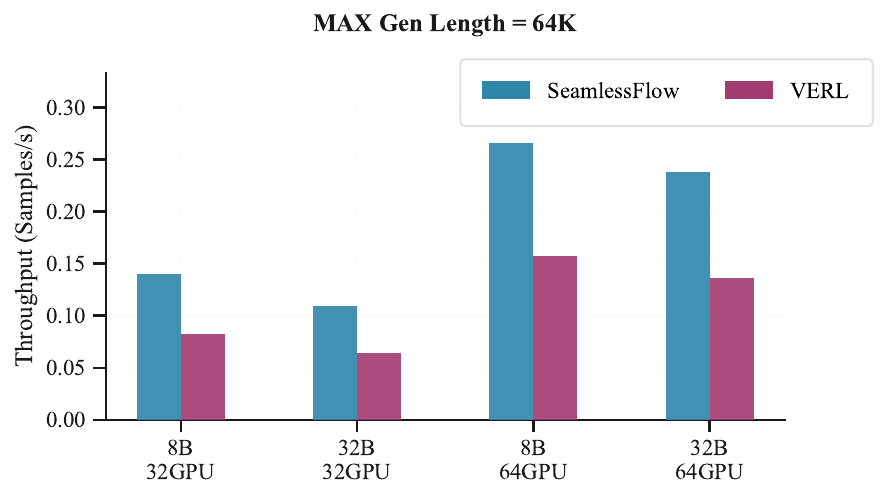}
  \caption{Throughput Comparison for Agentic Training.}
  \label{fig:mutil-tune}
\end{minipage}
\begin{minipage}[b]{0.48\textwidth}
  \centering
  \includegraphics[width=\linewidth]{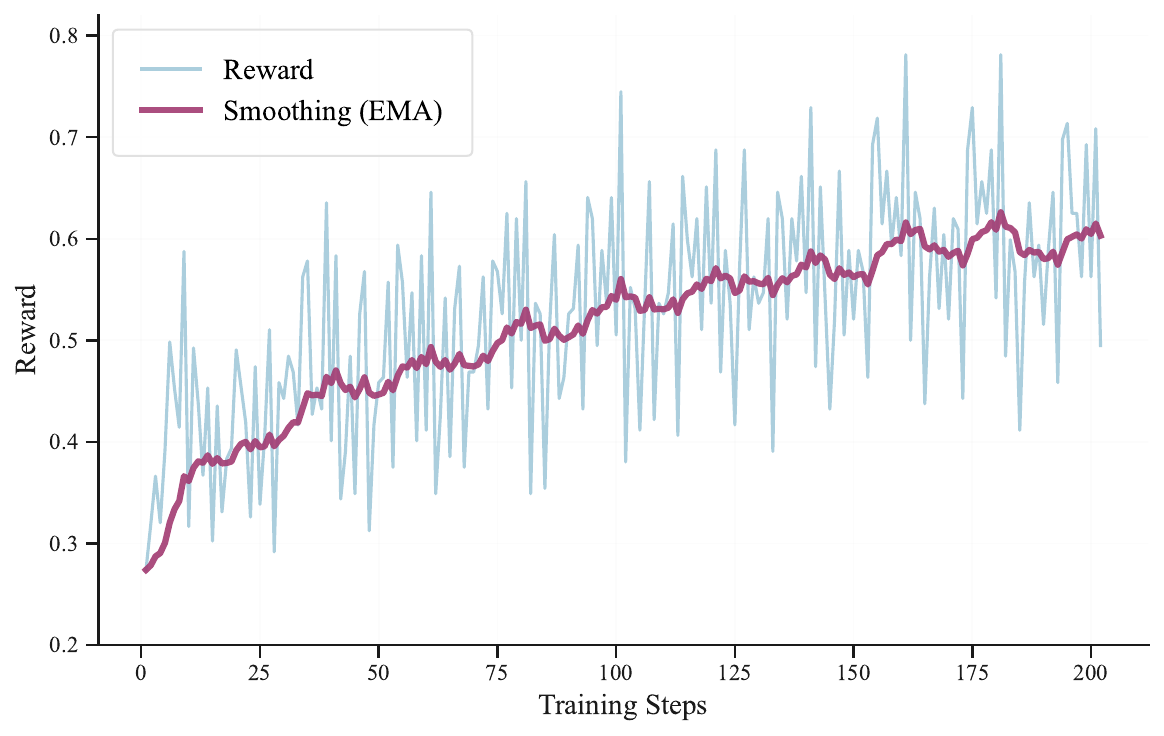}
  \caption{Reward Improvement Trajectories of Qwen3-32B During SeamlessFlow Training.}
  \label{fig:reward}
\end{minipage}\hfill
\end{figure}

For Agentic RL experiments, we use SWE-agent\cite{yang2024swe} as the scaffold to compare the performance of SeamlessFlow and VERL in agentic coding tasks. Using Qwen3\cite{yang2025qwen3}-32B and Qwen3-8B with a maximum generation length of 64K tokens, we perform comparative experiments on both 32-GPU and 64-GPU clusters. As illustrated in Figure \ref{fig:mutil-tune}, SeamlessFlow achieved an average throughput gain of 1.55x. Notably, the performance advantage of SeamlessFlow became more pronounced in the 64-GPU cluster setting, demonstrating its superior scalability in larger distributed training environments.

\subsection{End-to-End Agent Learning on Software Engineering Tasks}
To evaluate the performance of the models after conducting RL training with SeamlessFlow in complex agentic scenarios, we conducted a large-scale end-to-end RL training in software engineering tasks and evaluated in SWE-Bench\cite{jimenez2024swe}. 

Again, we use SWE-agent as our scafflod, and use Qwen3-32B as our base model. The training set contains 10K samples based on real issues from GitHub open-source code repositories, covering 3,500 different repositories and various types of programming tasks. Each data point includes complete problem descriptions with corresponding functional and regression test cases. Given the specificity and verifiability of coding tasks, we use test cases' pass-rate as reward signals.

As shown in Figure \ref{fig:reward}, the model has stable reward improvements during training. On the SWE-Bench Verified benchmark,  as shown in Table \ref{tab:swescore}, SeamlessFlow shows consistent improvements: Qwen3-8B improved from 12.2\% to 27.4\%, while Qwen3-32B achieved a more substantial gain from 23\% to 45.8\%.

\begin{table}[htbp]
\centering
\caption{Model Performance Trained with SeamlessFlow Framework}
\begin{tabular}{lc}
\toprule
Model& SWE-Bench Verified (\%)\\
\midrule
Qwen3-8B& 12.2\%\\
 Qwen3-8B-SeamlessFlow& 27.4\%\\
 Qwen3-32B& 23.0\%\\
Qwen3-32B-SeamlessFlow& 45.8\%\\ \bottomrule
\end{tabular}
\label{tab:swescore}
\end{table}

\section{Discussion}
RL training is now a key method for enhancing large language models (LLMs) and aligning them with human goals. Many RL frameworks have emerged to make this training faster, more scalable, and easier to deploy. We compare SeamlessFlow with several representative RL frameworks, highlighting the advantages it offers in addressing key challenges of large-scale RL.

Purely colocated designs such as VERL\cite{sheng2025hybridflow} cannot adapt to more general and large-scale RL tasks.
One reason is that they cannot leverage heterogeneous compute resources effectively.
Another limitation is that, in a colocated setup, LLM inference services must be suspended during training.
This makes it impossible to support scenarios like online RL in production environments,
where it is unacceptable for the serving service to be entirely unavailable for any period of time.
Compared to VERL, SeamlessFlow’s design can also operate in a purely colocated setup by simply assigning the same label to all GPU resources.
Beyond this, SeamlessFlow’s server-based, layered architecture is more modular and scalable than VERL’s single-controller design.
Furthermore, SeamlessFlow supports disaggregated setups, enabling better support for online RL scenarios where continuous serving must be maintained.

In disaggregated architectures, StreamRL\cite{zhong2025streamrl} and AsyncFlow\cite{han2025asyncflow} attempt to mask pipeline bubbles by using off-policy data delayed by one step. However, when there is a large runtime gap between rollout and training, delaying by one step is insufficient to eliminate most idle time. AReaL\cite{fu2025areal} mitigates bubbles by delaying even more steps, but at the cost of introducing large amounts of off-policy data that require special handling. In contrast, SeamlessFlow introduces spatiotemporal multiplexing: when training nodes are idle, their tasks are switched to rollout, naturally compressing training cluster idle time to the minimum and effectively eliminating pipeline bubbles.

For the isolation between the trainer and agents, contemporary work Agent Lightning\cite{luo2025agent} implements a similar separation server set between the trainer and agents. Compared to it, SeamlessFlow further handles model switching on the serving side in a way that is transparent to agents, achieving a more thorough decoupling. In addition, SeamlessFlow reduces the storage burden on the separation server through a more space-efficient trajectory merging design.

\section{Conclusion}
We presented SeamlessFlow, a server-based RL framework for stable, high-throughput training in heterogeneous industrial clusters.
SeamlessFlow’s agent-agnostic data plane decouples training from agent logic while guaranteeing bit-level trajectory consistency via proxy-managed capture, enabling immediate integration of new or modified agents.
Its tag-driven scheduling unifies colocated and disaggregated designs, dynamically reallocating resources to eliminate pipeline bubbles and maximize heterogeneous hardware utilization.
Together, these designs deliver stable, scalable RL for multi-agent, long-horizon deployments.
\newpage
\bibliography{main}
\newpage
\appendix

\end{document}